%% file: main.tex
\documentclass[10pt,twocolumn,letterpaper]{article}

\input{preamble}

\def\cvprPaperID{9899} 
\def\confName{CVPR}
\def\confYear{2022}

\begin{document}

\title{\vspace{-10mm}
Practical Imitation Learning in the Real World via Task Consistency Loss
}

\author{Mohi Khansari$^{1,*}$, Daniel Ho$^{1,*}$, Yuqing Du$^{1,2,*}$, Armando Fuentes$^1$, Matthew Bennice$^1$,\\ Nicolas Sievers$^1$, Sean Kirmani$^1$, Yunfei Bai$^1$, Eric Jang$^3$\\
{\it $^1$X, The Moonshot Factory, $^2$UC Berkeley, $^3$Google, $^*$Major contributors}\\
{\tt\small \{khansari, danielho, yuqingdu, mbennice, nsievers, skirmani, yunfeibai\}@x.team}\\
{\tt\small \{armandofuentes, ejang\}@google.com}\\
}
\maketitle

\input{abstract}
\input{introduction}
\input{related_work}

\input{problem_setup}
\input{method}
\input{experiments}
\input{results}
\input{conclusion}

{\small
\bibliographystyle{ieee_fullname}

\input{references.bbl}
}

\input{appendix.tex}

\end{document}

%% file: preamble.tex
\usepackage[pagenumbers]{cvpr} 

\usepackage{graphicx}
\usepackage{amsmath}
\usepackage{amssymb}
\usepackage{booktabs}
\usepackage{xcolor}
\usepackage{multirow}
\usepackage{wasysym}
\usepackage{float}
\usepackage[T1]{fontenc}
\usepackage[pagebackref,breaklinks,colorlinks]{hyperref}

\usepackage[capitalize]{cleveref}
\crefname{section}{Sec.}{Secs.}
\Crefname{section}{Section}{Sections}
\Crefname{table}{Table}{Tables}
\crefname{table}{Tab.}{Tabs.}

\newcommand{\upt}[1]{#1}
\newcommand{\overallsuccess}{\upt{72\%}~} 

%% file: abstract.tex
\begin{abstract}
Recent work in visual end-to-end learning for robotics has shown the promise of imitation learning across a variety of tasks. Such approaches are expensive both because they require large amounts of real world training demonstrations and because identifying the best model to deploy in the real world requires time-consuming real-world evaluations. These challenges can be mitigated by simulation: by supplementing real world data with simulated demonstrations and using simulated evaluations to identify high performing policies. However, this introduces the well-known ``reality gap'' problem, where simulator inaccuracies decorrelate performance in simulation from that of reality. In this paper, we build on top of prior work in GAN-based domain adaptation and introduce the notion of a \textit{Task Consistency Loss (TCL)}, a self-supervised loss that encourages sim and real alignment both at the feature and action-prediction levels. We demonstrate the effectiveness of our approach by teaching a mobile manipulator to autonomously approach a door, turn the handle to open the door, and enter the room. The policy performs control from RGB and depth images and generalizes to doors not encountered in training data. We achieve \overallsuccess success across sixteen seen and unseen scenes using only ${\sim}16.2$ hours of teleoperated demonstrations in sim and real. To the best of our knowledge, this is the first work to tackle latched door opening from a purely end-to-end learning approach, where the task of navigation and manipulation are jointly modeled by a single neural network.\end{abstract}

%% file: introduction.tex
\section{Introduction}
\label{sec:intro}

\begin{figure}
    \centering
    \includegraphics[width=.9\linewidth]{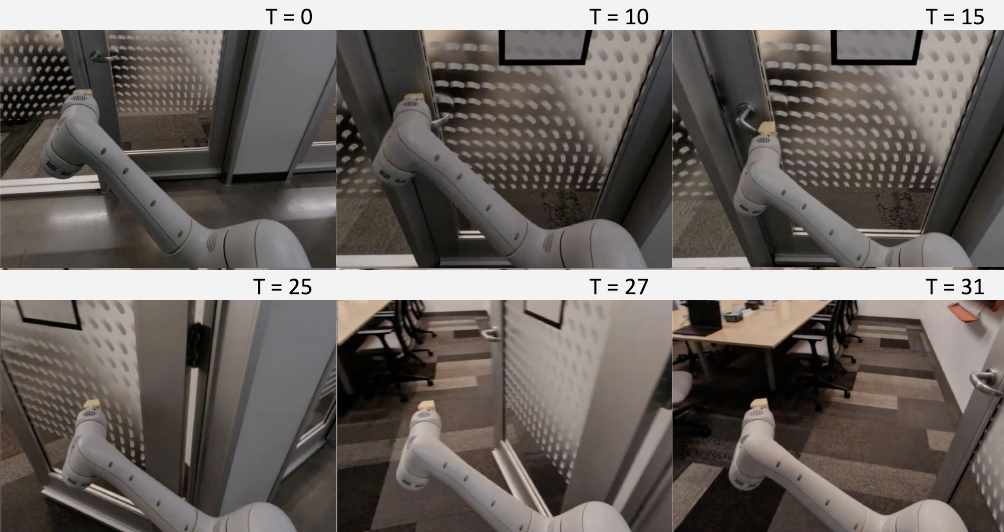}
    \vspace{-2mm}
    \caption{A sample door opening trajectory in a real world office environment using our method. The robot navigates to the door from 0-10s, unlatches the door from 10-20s, then fully opens the door and enters the room from 20-31s. }
    \label{fig:trajectory}
\end{figure}

\begin{figure}[ht]
    \centering
    \includegraphics[width=\linewidth]{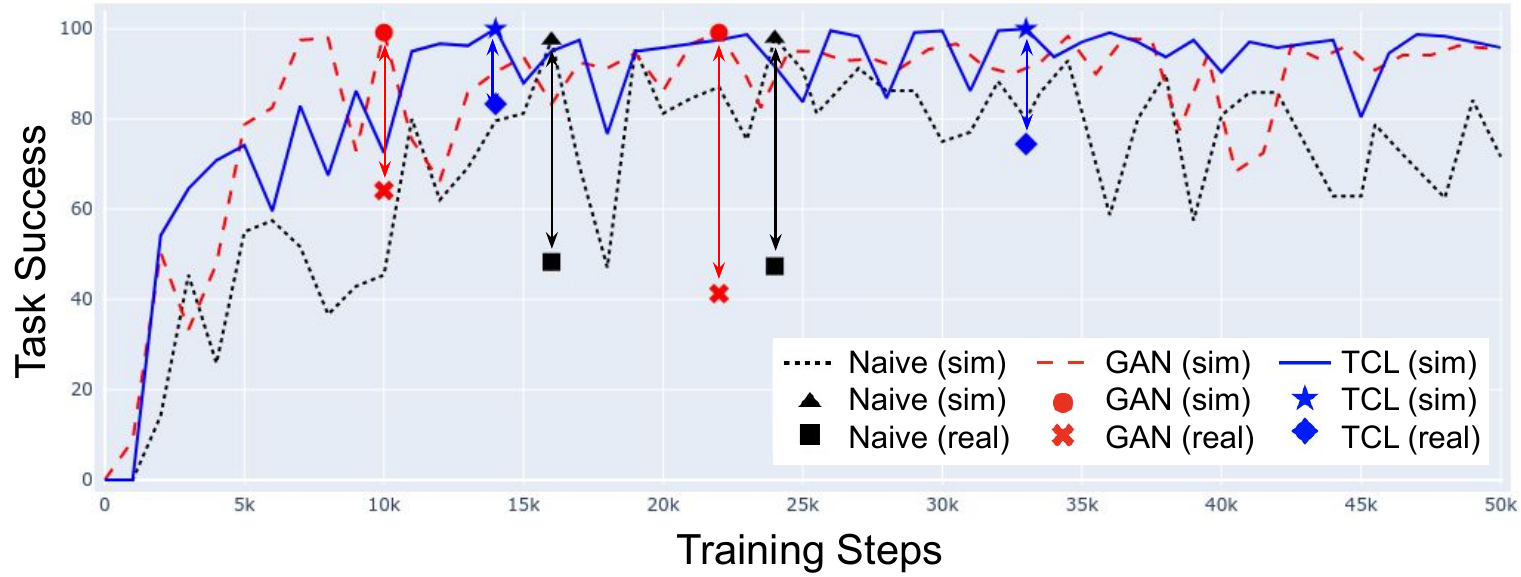}
    \vspace{-6mm}
    \caption{Matching sim and real evaluation performance is crucial to incrementally improving real-world performance in a cost-effective manner. Baseline methods of mixing sim + real data (Naive) and sim + real + GAN-adapted sim data (GAN) experience \upt{$49.9\%$} and \upt{$46.4\%$} performance drops due to reality gap. Our method, TCL, outperforms these baselines by reducing the gap to \upt{$21.1\%$}. All the three methods use RGB image as input.}
    \label{fig:success_compare_sim_real}
\end{figure}

\begin{figure*}[ht]
    \centering
    \includegraphics[width=\linewidth]{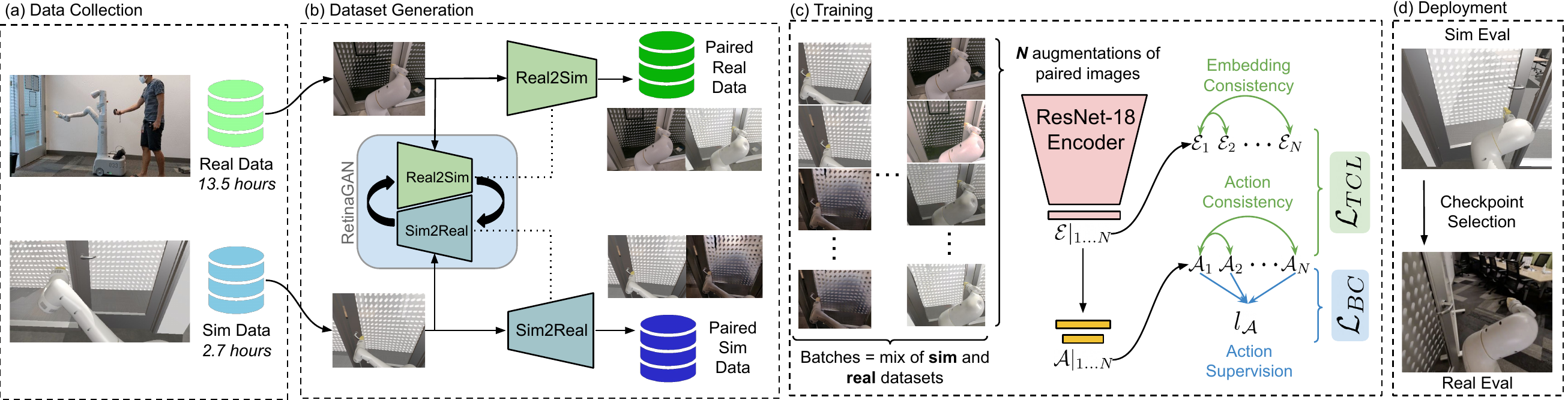}
    \vspace{-7mm}
    \caption{\textbf{(a, b)} We collect a set of sim and real images through teleoperation, and use them first to train a RetinaGAN model. \textbf{(b)} We then use the trained real-to-sim and sim-to-real models to create paired (real, adapted real) images and (paired sim, adapted sim) images, respectively. \textbf{(c)} Policy representations and predictions are encouraged to be invariant between paired images via a novel Task Consistency Loss. The same procedure can be used for the depth images (not shown here but used in the paper). \textbf{(d)} We use simulation in parallel to model training the evaluate all the checkpoints, and use it as a metric to select the right checkpoint for the real world deployment.}
    \label{fig:data_pipeline}
    \vspace{-2mm}
\end{figure*}

In recent years, the field of vision-based robotics has seen significant developments in navigation \cite{desouza2002vision, bonin2008visual, zhu2017target} or manipulation \cite{action_image, levine2016end} separately. However, if we eventually seek to deploy robots in human environments, we require agents capable of doing both simultaneously \cite{li2019hrl4in, wang2020learning}. 
Most prior work in vision-based manipulation focuses on fixed scenes from a third person perspective, but mobile manipulation introduces the challenge of precisely coordinating base and arm motions. Furthermore, manipulating objects from egocentric vision necessitates generalization to much greater visual diversity, since the robot's view is continuously changing as it moves through the environment. 

We choose to tackle this problem with imitation learning (IL), as recent work on end-to-end learning for manipulation has shown promising results with this approach \cite{rahmatizadeh2018vision, zhang2018deep, jang2021bcz}. However, imitation learning from raw sensor outputs requires numerous real world demonstrations. These demonstrations can be expensive and time consuming to collect, especially with the more complex action space of a mobile manipulator. 
Even after acquiring this data, evaluating learned policies in reality for generalization across a wide variety of unseen situations can still be time-consuming and hazardous. Unlike perception benchmarks, where validation datasets inform model selection, error on offline expert trajectories in robotics does not necessarily inform how the policy will behave if it drifts away from expert trajectories.

Simulators are often used to alleviate challenges with data collection and evaluation. For example, simulated demonstrations may be easier and safer to script and collect. The sim-to-real community often focuses on the ability to generate plentiful training data in simulation, but we posit that gathering enough real data to learn good policies is not too difficult; what is often far more time-consuming are the number of real-world trials needed to accurately compare policies across a number of generalization settings. Policies trained and evaluated in simulation suffer from the well known ``reality gap'', where visual and physical inaccuracies in the simulator can cause a high performing policy in simulation to still under-perform in the real world (see Figure \ref{fig:success_compare_sim_real}). In order to scale robotics to many real-world scenarios, we require a reliable simulated evaluation that is representative of real-world performance.

One popular and simple approach to bridging the reality gap is ``domain randomization'' \cite{tobin2017domain, sadeghi2017cadrl}, where a known set of simulator parameters, such as object textures and joint stiffness coefficients, are randomized within a hand-engineered range. Sufficient randomization will lead to a learned policy being robust to the true parameter values. 
Another approach is ``domain adaptation'', where the goal is to learn features and predictions invariant to the domain of model inputs. We build on past work in CycleGAN-based domain adaptation \cite{retinagan} by introducing additional feature-level and prediction-level alignment losses, the \textit{Task Consistency Loss}, between the adapted sim-to-real and real-to-sim images. We also extend our domain adaptation approach to the depth modality, showing our method can work with RGB, depth, and RGB-D inputs. Thus we leverage observations collected in both sim and reality for not just IL, but also for domain adaptation.

To test our approach, we focus on a challenging mobile manipulation task: \textit{latched door opening}. A mobile manipulator robot with head-mounted RGB-D sensors must autonomously approach a door, use the arm to turn the door handle, push the door open, and enter the room (Figure~\ref{fig:trajectory}). Prior work on door opening decouples the manipulation behavior from the navigation behavior, by first localizing the handle, planning an approach, then executing a grasping primitive \cite{stuede2019door}. In contrast, our method solely uses egocentric RGB-D images from the camera on the robot head and a single neural network for coordinating both arm and base motion to successfully open a variety of doors in an office building. In this paper, we will present an imitation learning system for mobile manipulation with a novel domain adaptation approach for aligning simulated and real performance. Our key contributions are:

\vspace{-3mm}
\begin{enumerate}
    \setlength\itemsep{-0.3em}
    \item To the best of our knowledge, this is the first work to tackle vision-based latched door opening with an end-to-end learning approach, encompassing: 1) navigation up to the door, 2) door unlatching and opening, and 3) entering the room. Our system generalizes to natural, unstructured human settings across a variety of time and lighting conditions. We achieve \overallsuccess success on 10 meeting rooms (6 seen and 4 doors during the training), with only $13.5$ hours of real demonstrations and $2.7$ hours of simulated demonstrations.
    \item Introducing feature-level and action-level sim and real alignment from a novel Task Consistency Loss, in addition to image-level alignment from modality-specific GANs. As shown in Figure~\ref{fig:success_compare_sim_real}, our method outperforms existing baselines of naively mixing real and sim and prior methods of GAN-adapted sim by a substantial margin of $+25$ percentage-point. 
\end{enumerate}

%% file: related_work.tex
\section{Related Work}
\label{sec:related}

\noindent \textbf{Deep Learning for Mobile Manipulation:} Although significant progress has been made in robot navigation and manipulation tasks individually, tackling the intersection of the two with deep learning is still relatively under-explored. Recent work has developed reinforcement learning methods for mobile manipulators, but are either only evaluated in simulation \cite{li2019hrl4in} or require many hours of real world learning \cite{gupta2018robot, sun2021fully}. The work by \cite{xia2021relmogen} proposes a hierarchical reinforcement learning approach for mobile manipulation tasks, but tackles a simpler variant of door opening, where the door opens by pushing a button or the door directly.
\cite{jang2021bcz} uses end-to-end imitation learning to push open swing doors (no handle) by driving the base of a mobile manipulator with the arm fixed. They improve performance in real by concatenating sim demonstrations and sim-to-real adapted images to the real demonstration dataset, but do not directly tackle the problem of narrowing the gap between simulated and real evaluation of the same model. We introduce a Task Consistency Loss to address that limitation, which enables us to scale end-to-end imitation learning to the harder task of latched door opening. 

A range of robotic control approaches have been proposed specifically for door opening, but require identifying the door handle through human intervention \cite{Jain2009BehaviorBasedDO} or additional sensor instrumentation \cite{petrovskaya2007probabilistic, schmid2008opening, peterson2000high, gu2017deep, welschehold2017learning}. For instance, \cite{stuede2019door} uses an object detector to identify the door handle and a scripted controller to grasp the handle to open the door. In contrast, our approach is fully end-to-end: navigation and manipulation decisions are inferred from first-person camera images without hand-engineering of object or task representations.

\noindent \textbf{Sim-to-real Transfer:} Prior work in sim-to-real transfer falls broadly in three categories: domain adaptation, domain randomization, and system identification. Our work focuses on domain adaptation, whereby discrepancies between sim and real are directly minimized. This could happen on the \textit{pixel}-level, where synthetic images are stylistically translated to appear more realistic, or on a \textit{feature}-level, where deep neural network features from simulation and real inputs are optimized to be similar.

Pixel-level domain adaptation work commonly make use of generative models to transfer inputs between domains, especially Generative Adversarial Networks (GANs) \cite{GAN}.
In robotics, this is frequently applied to robotic manipulation and grasping \cite{graspgan, rcan}.
Among these, RetinaGAN \cite{retinagan} translates images using perception-consistency to preserve object semantics and structure inherently important for robotic manipulation tasks.
RL-CycleGAN \cite{rlcyclegan} trains CycleGAN \cite{cyclegan} jointly with a reinforcement learning (RL) model. Here, consistency of RL predictions before and after GAN adaptation preserves visual qualities deemed important to RL learning. Our work also uses a notion of consistency; however, we apply it in the IL setting and aim instead to align domain representations with the goal of reducing the burden of checkpoint selection for deployment.

Feature-level domain adaptation work commonly analyze the \textit{distribution} of features from sim and real domains at the batch-level. DANN and DSN \cite{DANN, dsn} adversarially teach a network to extract features which does not discriminate between sim and real domains. Our feature-level domain adaptation method falls under self-supervised representation learning, which is commonly faciliated by increasing similarity between embeddings of positive image pairs. Prior work in this area has proposed using pairs generated from augmentations (e.g. random crop, flip, patch, colour shift) \cite{chen2020simple, henaff2020dataefficient, chen2020exploring}. We extend this approach to aligning paired simulated and real images from pixel-level domain adaptation GANs. That is, we maximize similarity between embeddings of the pairs (original sim, adapted sim) and (original real, adapted real).

Beyond embeddings, some approaches have posed classification or prediction self-supervision tasks using image context and invariants \cite{pathak2016context, mundhenk2018improvements, noroozi2016unsupervised, zhang2017split}. As image labels are invariant to augmentation, some methods aim to generalize or improve learning by learning augmentation strategies \cite{autoaugment, pba, cutout}. GAN adaptation could be considered a powerful learned augmentation adjusting the image domain.

Sim-to-real methods are utilized in mediated perception tasks in robotics, such as segmentation for autonomous driving \cite{wenzel2018modular} or pose estimation for object manipulation \cite{hodavn2020bop}. Because these tasks decouple perception from control, performance on real data are cheaply evaluated via metrics like IoU and AUROC on offline real data. However, evaluating end-to-end robot policies cannot be trivially done offline, and thus requires running multi-step predictions in the real world due to the causality effects (the current action can affect future observations, and future observations can further affect the proceeding actions). While our method can help with leveraging the simulation data for policy training similarly to previous domain adaptation works, it is additionally designed to help mitigate the cost of expensive real-world evaluation for end-to-end policies. One desideratum of our method is that simulated evaluation performance corresponds tightly to real world performance, and that this is achieved without much real-world tuning.

\noindent \textbf{Multimodal Learning:} Prior work in manipulation policies often use the RGB image alone as input. More recently, there's been a movement to use other modalities---such as depth, optical flow, and semantic segmentation \cite{zhao2019does,flowrl,fang2018multi,yan2018learning,yan2019data}---to improve sample efficiency and final performance of manipulation policies. While these derived higher-level modalities can implicitly be learned from the RGB image alone, using these geometric, semantic, and motion cues can improve training speed and task performance without the burden of learning from scratch.

%% file: problem_setup.tex
\section{Problem Setup}
\label{sec:problem_setup}

\subsection{Imitation Learning}

Our goal is to learn a policy, $\pi(a|s)$, that outputs a continuous action $a \in \mathcal{A}$ given an image $s \in \mathcal{S}$ which may be RGB, depth, or both. In imitation learning, we assume we have a dataset of expert demonstrations $\tau^* = (s_0, a_0, s_1, a_1, ... , s_{T-1}, a_{t-1}, s_T)$ with the actions generated by an expert policy $\pi^*$. We then learn to imitate this dataset with behaviour cloning, where the objective is to minimize a divergence between $\pi(a|s)$ and $\pi^*(a|s)$ given the same state $s$. Common minimization objectives are negative log-likelihood or mean-squared error. 

\subsection{Task}
We consider the task of latched door opening in a real office environment, in which the robot needs to drive a distance of ${\sim}1m$ to bring the arm in close vicinity of the door handle, use the arm to rotate the handle, and then use coordinated base and arm motions to swing the door open. This task has the following challenges: 

\vspace{-2mm}
\begin{enumerate}
\setlength\itemsep{-0.2em}
\item \textbf{High dimensional action space:} The task is only feasible by moving both the robot base (2-DoF) and the arm (7-DoF). A 9-dimensional action space together with high-dimensional visual inputs make this task particularly challenging for imitation learning, especially with a limited number of expert demonstrations.
\item \textbf{Mobile manipulation coordination:} The task requires precise coordination and time-synchronization between base and arm movements. For instance, there is no use in moving the arm if the handle is outside the robot's reachable space, and driving the base forward into a latched door leads to collision and robot arm breakage.
\item \textbf{Long horizon}: The task takes an expert $17$ to $60$ seconds to demonstrate, corresponding to up to 600 (input, action) pairs per episode. This long duration heightens task difficulty due to compounding errors associated with behavior cloning models \cite{ross2011reduction}.
\item \textbf{Bi-modal task nature}: We are training a single model to open both left-swing and right-swing doors, so the policy needs to infer the door swing direction and handle location from the image (See Figure \ref{fig:real_meeting_rooms}).
\end{enumerate}

\subsection{Data Collection}
We collect expert actions via teleoperation at 10Hz and record the corresponding RGB and depth image inputs. During the demonstration, the user can control both the robot base and arm via two handheld devices. We use the joystick on the left-hand device to command the base while using the 3D pose of the right-hand device to freely move the arm end-effector in the 3D space. 
\subsubsection{Real Dataset}
\vspace{-2mm}
\label{sec:problem_setup_protocol}
In total, we collected $2068$ real world demonstrations (corresponding to ${\sim}13.5$ hours) across 6 meeting rooms (3 left-swing and 3 right-swing doors). For each episode, we position the robot in front of the meeting room ${\sim}1$ meter away from the door. We then randomize the initial pose $\delta x \sim U(-0.25, +0.25)$ meters, $\delta y \sim U(-0.1, +0.1)$ meters, and $\delta \psi \sim U(-5, +5)$ degrees, where $x$ and $y$ correspond to the axes orthogonal and parallel to the door respectively, $\psi$ is the base orientation, and $U$ is the uniform distribution function. After initial pose randomization, we move the arm to a predefined initial joint configuration using the robot's built-in controller. We use a different initial configuration for the left and right swing doors to make the task more kinematically tractable. This prior knowledge of swing direction used in setup is not passed to the model; hence the model has to infer this from images.

After initial setup, the expert commands the robot via a hand-held teleoperation device and completes the episode when the door is sufficiently open such that the robot can enter the room without collision. We do not control the condition of the room (light, chair, table, ...) and collect demonstrations in the natural state left by previous users.
\vspace{-2mm}

\subsubsection{Sim Dataset}
\begin{figure}[]
    \centering
    \includegraphics[width=\linewidth]{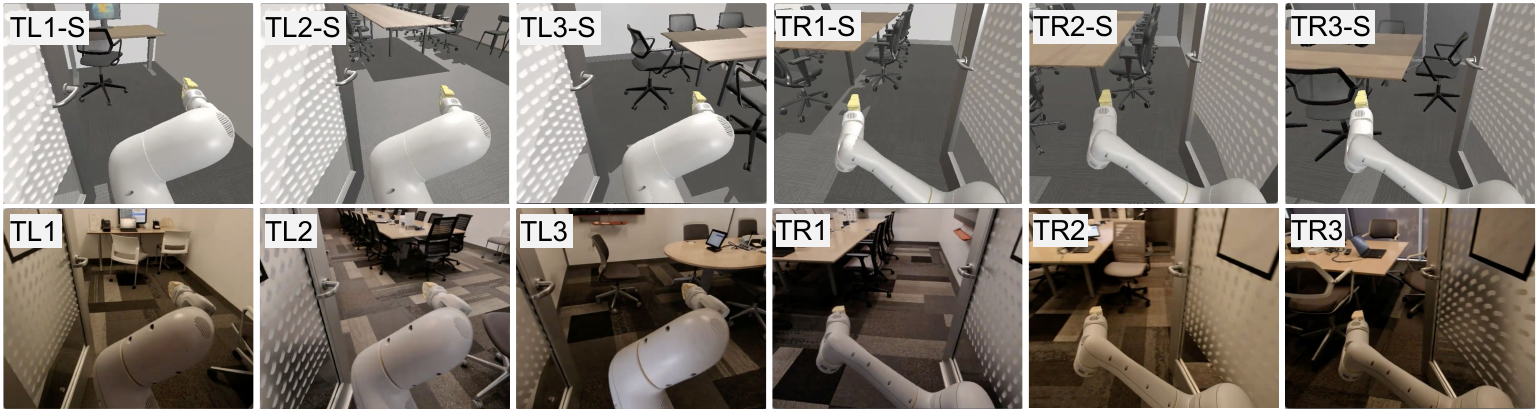}
    \vspace{-7mm}
    \caption{\textbf{Above:} sim scenes of the 6 training meeting rooms, used for sim teleoperation data collection and checkpoint evaluations. \textbf{Below:} real meeting rooms. The assigned name for each room is indicated on the top-left corner and are created based on Table \ref{tab:eval_protocol}.}
    \vspace{-2mm}
    \label{fig:sim_rooms}
\end{figure}

\vspace{-2mm}
We create 3D models of the 6 training meeting rooms with lower-fidelity textures but sufficient structural detail for
the RetinaGAN domain adaptation model to translate to real (see Figure \ref{fig:sim_rooms}). During sim data collection, we use the same teleoperation interface, task setup, and success metric as in real. In total, we collected ${\sim}500$ demonstrations, corresponding to ${\sim}2.7$ hours of data.

%% file: method.tex
\section{Method}
\label{sec:method}

Our method leverages the domain adaptation GAN works, RetinaGAN \cite{retinagan} and CycleGAN \cite{cyclegan}, and extends them by further reducing the sim-to-real gap not only at the visual level, but also at the feature and action prediction level using the Task Consistency Loss (TCL). We use the following notation:

\vspace{-3mm}
\begin{itemize}
    \setlength\itemsep{-0.3em}
    \item Subscripts $_{RGB}$ and $_{D}$ reference parameters or functions associated with RGB and depth images, respectively.
    \item $\mathcal{I}$ refers to an input image, either RGB, $\mathcal{I}_{RGB} \in \mathcal{R}_+^{H \times W \times 3}$, or Depth, $\mathcal{I}_{D} \in \mathcal{R}_+^{H \times W}$.
    \item $\mathcal{D}$ references an image augmentation/distortion function. For RGB, $\mathcal{D}_{RGB}$, we apply random crop, brightness, saturation, hue, contrast, cutout, and additive Gaussian noise. For depth, $\mathcal{D}_{D}$, we only apply random crop and cutout.
    \item $\mathcal{G}$ refers to sim2real $\mathcal{G}^{sim2real}$ or real2sim $\mathcal{G}^{real2sim}$ generators of RetinaGAN or CycleGAN models. We use separate GANs for each modality. For example, $\mathcal{G}^{sim2real}_{RGB}$ transfers RGB images from the sim domain to the real domain.
\end{itemize}

\vspace{-2mm}
For brevity, we may drop subscripts and superscripts to indicate that a process can be applied on either input modality. For instance, $\mathcal{I}$ indicates use of either RGB or depth images. Examples of transformed RGB and depth images through $\mathcal{D}$ and $\mathcal{G}^{sim2real}$ are shown in Figure \ref{fig:pairs}.

\subsection{Paired Image Generation using GANs}
\label{sec:gan}
We visually align images from unpaired sim and real datasets by building on top of the pixel-level domain adaptation techniques, RetinaGAN \cite{retinagan} and CycleGAN \cite{cyclegan}, by extending them to the latched door opening task. From these models, we use the sim2real and real2sim generator networks to adapt images from our original demonstrations. The resulting datasets contain an original sim or real image and the corresponding domain-translated \textit{paired} image. 

\noindent \textbf{RGB GAN}:
We train a GAN using the perception consistency loss based on Section V.C of the RetinaGAN work \cite{retinagan}, re-using the off-the-shelf RetinaNet object detector trained on object grasping examples \cite{retinanet}. RetinaGAN trains unsupervised, using only images collected from teleoperation, described in Section \ref{sec:problem_setup}. Within GAN-translated RGB images of simulation, glass door patterns appear more translucent, lighting conditions more randomized, lighting effects like global illumination and ambient occlusion added, and color tones adjusted. This process is reversed in GAN-translated real images.

\begin{figure}[t!]
    \centering
    \includegraphics[width=\linewidth]{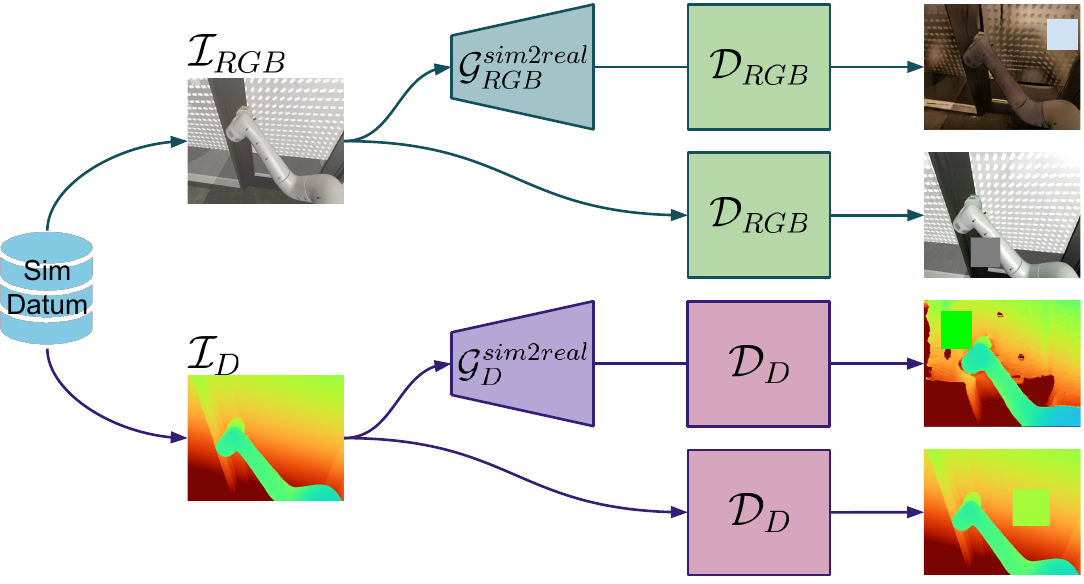}
    \vspace{-7mm}
    \caption{Illustration of applying augmentations through $\mathcal{D}$ and $\mathcal{G}^{sim2real}$ to an input dataset from simulation. Turbo colormap applied to depth images for clarity. This process is reversed in GAN-translated real images (not shown in here).}
    \label{fig:pairs}
\end{figure}

\noindent \textbf{Depth GAN}: 
For the depth modality, we train a CycleGAN \cite{cyclegan} model---we lack a depth detector needed for RetinaGAN---on stereo real depth (computed using HitNet \cite{hitnet} stereo matching) and simulated ground truth depth images. We pre-process images by clipping depth to 10 meters.
The trained model reliably translates between differences in the two domains. Foremost, real images have significant noise from sensors and stereo matching, while simulation images are noiseless. The glass and privacy film of the doors appear as opaque in simulation but translucent in real, where depth bleeds through to the floor of the conference room behind. The depth GAN learns to inpaint real image pixels which have passed through the door, and it generates patches of depth behind the glass in simulation images.
Figure \ref{fig:pairs} shows an example of adapted sim images.

\subsection{Task Consistency Loss (TCL)}
\label{sec:tcl}

In addition to adaptation at the pixel level through GANs, we introduce a novel auxiliary loss, TCL, to encourage stronger alignment between the sim and real domains for adaptation at the feature and the action-prediction levels. 
For a given image $\mathcal{I}$, we can generate $N$ variations, $\mathcal{I}|_{1..N}$, by applying augmentations such as $\mathcal{D}$, $\mathcal{G}$, or both.
In this paper we consider the following three variations for an input image $\mathcal{I}$:

\vspace{-2mm}
\begin{itemize}
    \setlength\itemsep{-0.3em}
    \item Original sim/real image distorted with $\mathcal{D}$,  $\mathcal{I}_1$ = $\mathcal{D} (\mathcal{I})$
    \item A distorted instance of the original sim/real image, $\mathcal{I}_2$ = $\mathcal{D} (\mathcal{I})$. The consistency loss between $\mathcal{I}_1$ and $\mathcal{I}_2$ enforces invariancy with respect to the applied image distortion transformations.
    \item Adapted original images via $\mathcal{G}$ followed by a distortion, $\mathcal{I}_3$ = $\mathcal{D} (\mathcal{G}(\mathcal{I}))$. The consistency loss between $\mathcal{I}_1$ and $\mathcal{I}_3$ enforces invariancy with respect to the domain transformation as well as the image distortions.
\end{itemize}

\begin{figure}[t]
    \centering
    \includegraphics[width=.9\linewidth]{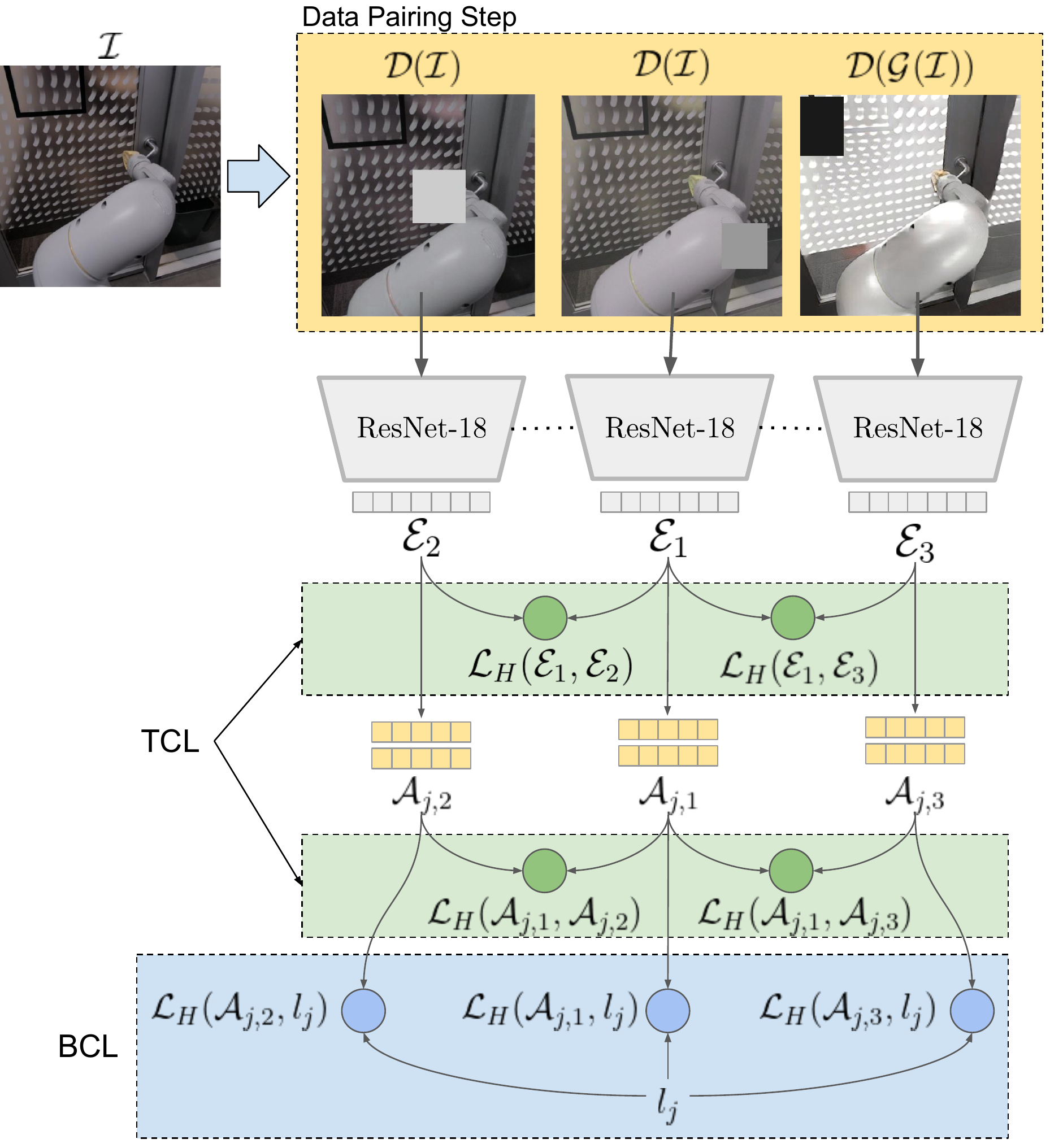}
    \vspace{-3mm}
    \caption{Task Consistency Loss. We create positive pairs by 1) augmenting the image, and 2) adapting the image from sim-to-real or real-to-sim with the corresponding GAN, then applying augmentations. We pass all images of the same modality through the same ResNet-18 \cite{he2015deep} encoder $f_{\phi}$ followed by a normalization layer to generate embeddings $\mathcal{E}_i$, and then pass them through a two layer MLP $g_{\phi}$ to get the predicted actions $\mathcal{A}_{j,i}$. Thus, for each image we can compute $\mathcal{L}_{TCL}$ and $\mathcal{L}_{BC}$, using $\mathcal{E}_i$, $\mathcal{A}_{j,i}$, $\forall i \in 1..N$ and $j \in (a, b, f)$, where $\mathcal{A}_b$, and $\mathcal{A}_f$ correspond to predicted actions for arm, base, and termination, respectively.}
    \vspace{-7mm}
    \label{fig:TCL}
\end{figure}

The $N$ variations of the input image $\mathcal{I}|_{1..N}$ depict the same instant of time. Hence, the image embeddings $\mathcal{E}|_{1..N}$ and predicted actions $\mathcal{A}|_{1..N}$ should be invariant under augmentations $\mathcal{D}$ and $\mathcal{G}$, and we derive our self-supervised signal by enforcing this invariancy. We hypothesize that this will help close the sim-to-real gap and make performance in simulation more representative of that in reality. Additionally, imposing this consistency loss on images augmented with random cutout may improve robustness to occlusions; it encourages the model to learn features in context of other salient features (e.g. the handle based on the door frame, see Figure \ref{fig:TCL}).

To calculate TCL, we pass all variations of the input image through the same network to calculate corresponding image embeddings $\mathcal{E}|_{1..N}$ and estimated actions $\mathcal{A}|_{1..N}$. Then, we apply a Huber loss $\mathcal{L}_{H}$ \cite{huber} to penalize discrepancies between pairs as follows:

\vspace{-5mm}
\begin{equation}
    \mathcal{L}_{TCL} = \sum_{i=2} ^ N  \Big( \mathcal{L}_{H} (\mathcal{E}_1, \mathcal{E}_i) + \sum_{j \in (a, b, f)} \mathcal{L}_{H} (\mathcal{A}_{j, 1}, \mathcal{A}_{j, i}) \Big)
\end{equation}
\vspace{-3mm}

\noindent where the first term imposes consistency loss over the embeddings and the second term penalizes estimated action errors between all  variations. Note that $\mathcal{A}_a$, $\mathcal{A}_b$, and $\mathcal{A}_f$ correspond to predicted actions for arm, base, and termination, respectively.
The augmentation and loss setup for the feature-level TCL is shown in Figure \ref{fig:TCL}.

\subsection{Behavior Cloning Loss (BCL)}\label{sec:bc}

The behavior cloning loss is applied at each network head to enforce similarity between predicted actions $\mathcal{A}_j$ and demonstrated labels $l_j$, $\forall j \in (a, b, f)$. We use the same label to calculate BCL for all $N$ variations of the input image, which can further reinforce invariancy across applied image augmentations:

\vspace{-3mm}
\begin{equation}
    \mathcal{L}_{BC} = \sum_{j \in (a, b, f)} \sum_{i=1} ^ N \mathcal{L}_{H} (\mathcal{A}_{j, i}, l_j)
\end{equation}

The overall policy training loss used is:

\vspace{-3mm}
\begin{equation}
    \mathcal{L} = \mathcal{L}_{BC} + \mathcal{L}_{TCL}
\end{equation}

\subsection{Multi-Sensor Network Architecture}
\begin{figure*}[ht]
    \centering
    \includegraphics[width=.8\linewidth]{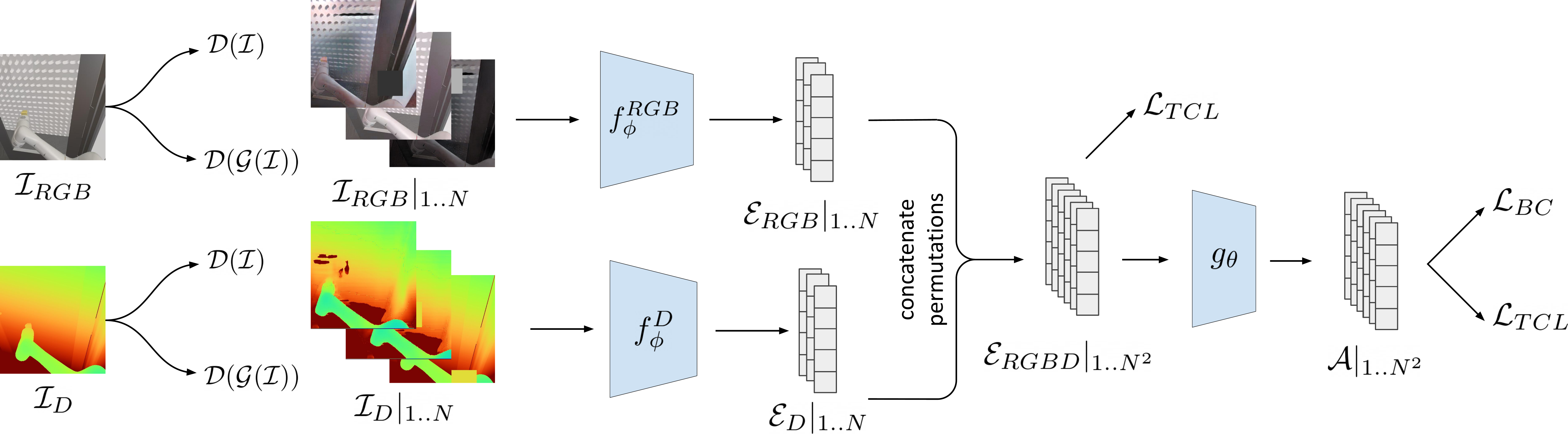}
    \vspace{-2mm}
    \caption{Multimodal network architecture. Given a frame $\mathcal{I} = (\mathcal{I}_{RGB}, \mathcal{I}_{D})$, we first apply augmentations and/or domain adaptations to create $N$ variants of $\mathcal{I}$. We then pass modalities through separate ResNet-18 \cite{he2015deep} encoders, $f_\phi^{RGB}, f_\phi^{D}$, producing $N$ embeddings per modality. We combine modalities by concatenating all permutations of embeddings, for $N^2$ total combinations. The $N^2$ combinations are then passed through a 2-layer MLP to get the predicted actions. We apply $\mathcal{L}_{TCL}$ to embeddings and actions and $\mathcal{L}_{BC}$ to actions. }
    \label{fig:multimodalsys}
\end{figure*}

The overall multi-sensor network is shown in Figure \ref{fig:multimodalsys}. We use the methods described in Section \ref{sec:gan} to generate domain adapted and augmented images for each modality, then apply TCL as described in Section \ref{sec:tcl}. To combine the different modalities, we concatenate all permutations of the $N$ different variations per modality to get $N^2$ RGB-D embeddings. Empirically, we find that sensor fusion at the embedding level leads to higher task success than channel-wise fusion of the raw RGB and depth images prior to passing to the ResNet-18 \cite{he2015deep} encoders. We then pass the concatenated embeddings through a fully connected network to compute action predictions for the BCL as described in Section \ref{sec:bc}.

%% file: experiments.tex
\section{Experiments}
\label{sec:experiments}

\subsection{Evaluation Protocol}
\begin{figure*}[h!]
    \centering
    \includegraphics[width=.9\linewidth]{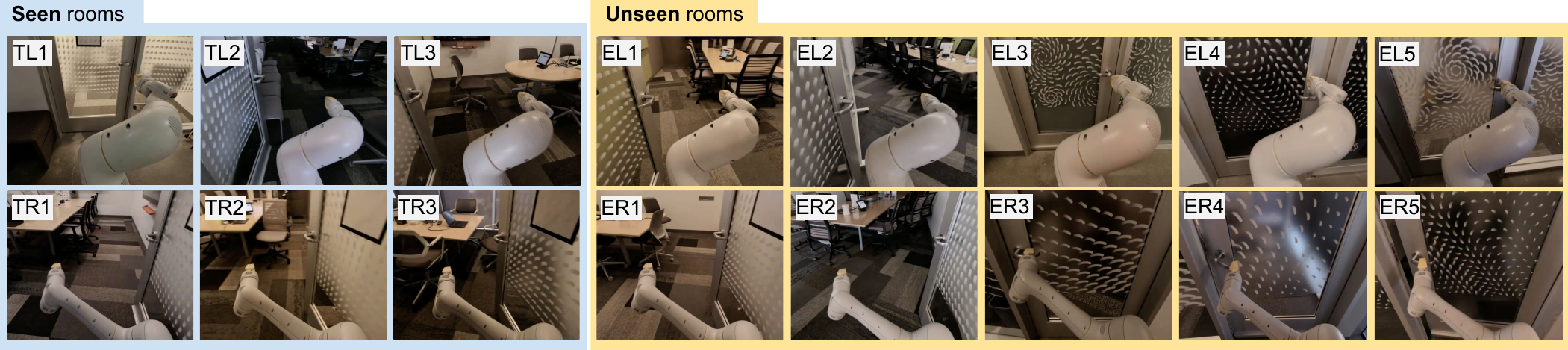}
    \vspace{-2mm}
    \caption{A snapshot from each of the sixteen meeting rooms used in our experiment, with room names at the top-left of each image according to the naming convention in Table \ref{tab:eval_protocol}. See TL2 for an example of camera image captured when the room's lights are off.}
    \label{fig:real_meeting_rooms}
\end{figure*}

\begin{table*}[h!]
  \centering
  \footnotesize{
  \begin{tabular}{l|ccc|ccc||ccccc|ccccc}
     &  \multicolumn{6}{c||}{Train} &  \multicolumn{10}{c}{Eval}\\ \hline 
    Swing &  \multicolumn{3}{c|}{Left} &  \multicolumn{3}{c||}{Right} &  \multicolumn{5}{c|}{Left} &  \multicolumn{5}{c}{Right} \\ \hline
    Bot A & \Circle \textcolor{orange}{\sun} &  \LEFTcircle \lightning &  \CIRCLE \textcolor{orange}{\sun} & \Circle \textcolor{orange}{\sun} &  \LEFTcircle \lightning &  \CIRCLE \lightning            &          \Circle \lightning & \Circle \textcolor{orange}{\sun} & \LEFTcircle \textcolor{orange}{\sun} & \CIRCLE \lightning & \CIRCLE \lightning           &          \Circle \textcolor{orange}{\sun} &  \LEFTcircle \lightning & \Circle \textcolor{orange}{\sun} & \Circle \textcolor{orange}{\sun} & \LEFTcircle \lightning     \\

    Bot B &  \LEFTcircle \lightning &  \CIRCLE \textcolor{orange}{\sun} & \Circle \lightning &  \LEFTcircle \lightning &  \CIRCLE \textcolor{orange}{\sun} & \Circle \textcolor{orange}{\sun}            &          \CIRCLE \textcolor{orange}{\sun} &  \LEFTcircle \lightning & \CIRCLE \lightning & \LEFTcircle \textcolor{orange}{\sun} & \Circle \textcolor{orange}{\sun}         &            \CIRCLE \lightning &  \CIRCLE \textcolor{orange}{\sun} & \CIRCLE \lightning & \LEFTcircle \lightning & \Circle \textcolor{orange}{\sun}      \\ \hline \hline
    Name & TL1 & TL2 & TL3 & TR1 & TR2 & TR3 & EL1 & EL2 & EL3 & EL 4 & EL5 & ER1 & ER2 & ER3 & ER4 & ER5
  \end{tabular}
  }
  \vspace{-3mm}
  \caption{Evaluation protocol: We test on 16 meeting rooms and two robots, at different times of day and lighting conditions. Train doors have corresponding demonstrations in the training set, while Eval doors are unseen during training. Room status is described with two symbols: the first symbol indicates the time of the day [\Circle \;Morning, \LEFTcircle \;Noon, \CIRCLE \;Afternoon], and the second symbol indicates lighting status [\textcolor{orange}{\sun} \;On, \lightning \;Off]. The last row shows the abbreviated naming convention for Figure \ref{fig:real_meeting_rooms}. }
  \label{tab:eval_protocol}
\end{table*}

We evaluate the performance of our model on 16 latched doors, with 6 doors for training (3 left swinging and 3 right swinging) and 10 solely for evaluation (5 left swinging and 5 right swinging) (see Figure \ref{fig:real_meeting_rooms}). For each door, we evaluate with 30 trials on two mobile manipulators, Robot A and Robot B, and only Robot A was used to collect training data. For consistency between evaluations across models, we split the time of evaluation between three categories: morning (8AM-11AM), noon (11AM-2PM), and afternoon (2PM-5PM) and ensured all models for each room are evaluated in the same time category. We shut the window blinds in all evaluations and controlled whether room lights were turned on. Table \ref{tab:eval_protocol} provides a summary of the evaluation protocol used for each room. As these rooms are also in use by others, the types of objects and poses of interior furniture were continuously changing during our multi-week evaluations.

We use the same initial setup as during data collection and follow the same guidelines to determine task success/failure (see Section \ref{sec:problem_setup_protocol}). After initial setup, the policy controls the robot autonomously to perform the task. The safety operator can intervene at any moment to stop the robot if needed, which automatically marks the particular evaluation as a failure. All models are trained to predict task termination based on the input images. A policy which does not terminate within a timeout of two minutes is also marked as a failure.

We consider two baseline approaches: 1) RGB-Naive Mixing: trained on naively mixing of sim and real images, 2) RGB-GAN \cite{retinagan}, trained on three sources of data: RGB sim images, RGB real images, and RGB sim images adapted using a sim2real GAN. Both of these are ablations of our method, with 1) ablating domain adaptation entirely and 2) ablating real2sim adaptation and TCL. 

We compare the baselines against three instances of our method: 1) RGB-TCL: An RGB-only model with TCL on the three variations of input images described in Section \ref{sec:tcl}, fed from both sim and real datasets, 2) Depth-TCL: Similar to (1), but with depth images as input, and 3) RGBD - TCL: A multi-sensor variant with both RGB and depth images as per Figure \ref{fig:multimodalsys}.

To account for variations in model training and create a fair comparison, we train three models for each approach with different random seeds and export new model checkpoints at 10 minute intervals. We use 250 simulation worker instances to evaluate the performance of each checkpoint in simulation. As described in Section \ref{sec:intro}, this thorough simulation evaluation is necessary to pick the right checkpoint; for imitation learning models, we cannot reliably determine when a model starts to overfit and then apply early stopping solely through the offline validation dataset.
Based on sim evaluations across ${\sim}300$ checkpoints and three models, we evaluate the top-three checkpoints in a blind real-world evaluation: checkpoints are chosen at random between episodes so operators do not know which models they evaluate.

%% file: results.tex
\subsection{Results}
\label{sec:results}

\begin{table}
  \centering
  \footnotesize{
  \begin{tabular}{@{}l|c|cc@{}}
    \toprule
    Method & Total & Seen & Unseen \\
    \midrule
    RGB - Naive (baseline) & 31\% $\pm$ 2.1 & 48\% $\pm$ 3.7 & 21\% $\pm$ 2.4\\
    RGB - GAN (baseline)   & 48\% $\pm$ 2.3 & 56\% $\pm$ 3.7 & 43\% $\pm$ 2.9\\
    RGB - TCL              & 63\% $\pm$ 2.2 & 74\% $\pm$ 3.3 & 56\% $\pm$ 2.9\\
    Depth - TCL    & \textbf{72\% $\pm$ 2.1}& \textbf{81\% $\pm$ 3.0} & \textbf{67\% $\pm$ 2.7}\\
    RGBD - TCL             & 69\% $\pm$ 2.1 & \textbf{80\% $\pm$ 3.0} & 63\% $\pm$ 2.8\\
    \bottomrule
  \end{tabular}
  }
  \vspace{-3mm}
  \caption{Door opening success rate (\%) $\pm$ standard deviation in \textbf{real}, using various imitation learning methods. The total performance is calculated based on 480 evals (180 from seen and 300 from unseen doors).}
  \label{tab:success_compare}
  \vspace{-2mm}
\end{table}

\begin{table*}
  \centering
  \footnotesize{
  \begin{tabular}{l||cc||cc||cc}
    \toprule
    Method &  \multicolumn{2}{c}{Swing Orientation} &  \multicolumn{2}{c}{Lighting} &  \multicolumn{2}{c}{Robot}  \\ \hline
     & Right & Left & On & Off & A & B  \\
    \hline
    RGB - Naive & 39\% $\pm$ 3.2 & 24\% $\pm$ 2.8  & 40\% $\pm$ 3.2 & 23\% $\pm$ 2.7  & 32\% $\pm$ 3.0 & 30\% $\pm$ 3.0 \\
    RGB - GAN   & 51\% $\pm$ 3.2 & 45\% $\pm$ 3.2  & 52\% $\pm$ 3.2 & 44\% $\pm$ 3.2  & 43\% $\pm$ 3.2 & 53\% $\pm$ 3.2 \\
    RGB - TCL   & 59\% $\pm$ 3.2 & 66\% $\pm$ 3.1  & 88\% $\pm$ 3.0 & 55\% $\pm$ 3.2  & 59\% $\pm$ 3.2 & 66\% $\pm$ 3.1 \\
    Depth - TCL & 71\% $\pm$ 2.9 & 73\% $\pm$ 2.9  & 76\% $\pm$ 2.9 & 73\% $\pm$ 2.9  & 72\% $\pm$ 2.9 & 72\% $\pm$ 2.9 \\
    RGBD - TCL  & 75\% $\pm$ 2.8 & 63\% $\pm$ 3.1  & 82\% $\pm$ 2.9 & 67\% $\pm$ 3.0  & 66\% $\pm$ 3.1 & 73\% $\pm$ 2.9 \\
    \bottomrule
  \end{tabular}
  }
  \vspace{-3mm}
  \caption{Breakdown of performance by door swing direction, lighting status, and robot variant (training data was only from A). Each result is calculated from 150 evaluations, but across 5 meeting rooms (3 seen and 2 unseen) for the swing orientation and across 10 meeting rooms (6 seen and 4 unseen) for the rest.}
  \vspace{-3mm}
  \label{tab:success_compare_methods}
\end{table*}

The experiment results on latched door opening success are provided in Table \ref{tab:success_compare}. We report estimated standard deviation for each experiment as $\sqrt{p(1-p)/(n-1)}$, assuming $n$ trials that are i.i.d. Bernoulli variables with success rate $p$. As expected, RGB-Naive has the worst performance of $31\%$ since there is no explicit forcing function to reduce the domain gap. Using the RetinaGAN sim-to-real model, RGB-GAN improves \upt{$17\%$} over the RGB-Naive model. Finally, by imposing the task consistency loss at both feature and action levels, the RGB-TCL model outperforms the RGB-Naive and RGB-GAN baselines by \upt{$32\%$} and \upt{$15\%$}, respectively. The Depth-TCL has the highest performance of \overallsuccess followed by RGBD-TCL with $69\%$. RGB-TCL, with $63\%$ success, has a lower performance than the other TCL variations, most likely due to higher impact of visual difference in the RGB channel.

Figure \ref{fig:success_compare_sim_real} further compares sim and real performance for one run of RGB-Naive, RGB-GAN, and RGB-TCL. We observe from the figure that: 
(a) Sim performance fluctuates for all methods as training progresses, despite validation losses (not shown) decreasing near monotonically. As a result, always selecting the last checkpoint or basing off of validation loss is not sufficient. 
(b) Variance across training steps is highest for RGB-Naive and lowest for RGB-TCL. Within RGB-Naive, we hypothesize that sim and real domains are encoded as separate features and converge separately w.r.t. task success. In contrast, RGB-TCL model encodes domain invariant features and is thus more stable. We plot real world performance of the top two checkpoints for each model and measure the average sim-real performance gap for RGB-Naive, RGB-GAN, and RGB-TCL as \upt{$49.9\%$}, \upt{$46.4\%$} and \upt{$21.1\%$}, respectively.

We would like to point out that each real world evaluation takes almost a full day to converge, in contrast to ${\sim}10$ minutes in simulation. This solidifies the importance of reliable simulation and sim-to-real transfer in guiding checkpoint selection for evaluation. 

Table \ref{tab:success_compare_methods} compares performance w.r.t. three factors: door swing direction, room light status, and the robot used. The Depth-TCL model has the least variations across all these three variations, and together with its highest overall performance, making it the most suitable mode of operations for the latched door opening task.

%% file: conclusion.tex
\section{Conclusion}
\label{sec:conclusion}
\vspace{-2mm}
In this work we presented the Task Consistency Loss (TCL), a self-supervised method for sim and real domain adaptation at the feature and action levels. Real world robotic policy evaluation for mobile manipulators can be laborious and hazardous. TCL allows us to leverage simulation to identify promising policies for real world deployment, while mitigating the reality gap. We demonstrated our method on latched door opening, a challenging mobile manipulation task, using only egocentric RGB-D camera images. With only 13.5 hours of real world demonstrations and 2.7 hours of simulated demonstrations, we showed that our method improves real world performance on both seen and unseen doors, reaching \overallsuccess success. We demonstrated that using TCL reduces the gap between sim and real model evaluations by $+24$ percentage-point relative to the baselines. This opens an opportunity to evaluate in sim to select more optimal models for real world deployment.

\input{discussion}

%% file: discussion.tex
\textbf{Limitations and Future Work:}
TCL helps mitigate the sim-to-real gap via TCL, but does not completely remove it. Section \ref{sec:results} shows that there is still a gap of \upt{$21.1\%$} between domains.
Furthermore, given that our approach uses the generators from RetinaGAN/CycleGAN in the dataset pairing process, selecting a poor generator can yield poor TCL performance. One mitigation is to randomly select amongst a pool of candidate checkpoints during data-pairing, to avoid locking in an unlucky checkpoint. We hypothesize that sampling random GAN checkpoints in conjunction with TCL makes the policy more robust, and is analogous to a rich data augmentation or domain randomization strategy, and aim to pursue this in future work.

\textbf{Potential Negative Societal Impacts:} Although our policy achieves high success rate, we caution that an explicit safety layer for human-robot and robot-environment interaction was not within the scope of this paper, and potential safety issues of mobile manipulation are greater than either navigation-only (e.g. unknown workspace, but no contacts) or manipulation-only research (e.g. contacts in a known workspace). One potential mitigation that does not compromise the end-to-end generality of our approach is to have the policy explicitly model safety-relevant predictions and decisions from a diverse dataset of human-robot and robot-environment interactions.

%% file: appendix.tex
\onecolumn
\section*{Appendix}
\appendix

\setcounter{section}{0}

\section{Network Architecture}
\label{appendix:networkarch}

\begin{figure*}[h]
    \centering
    \includegraphics[width=0.9\textwidth]{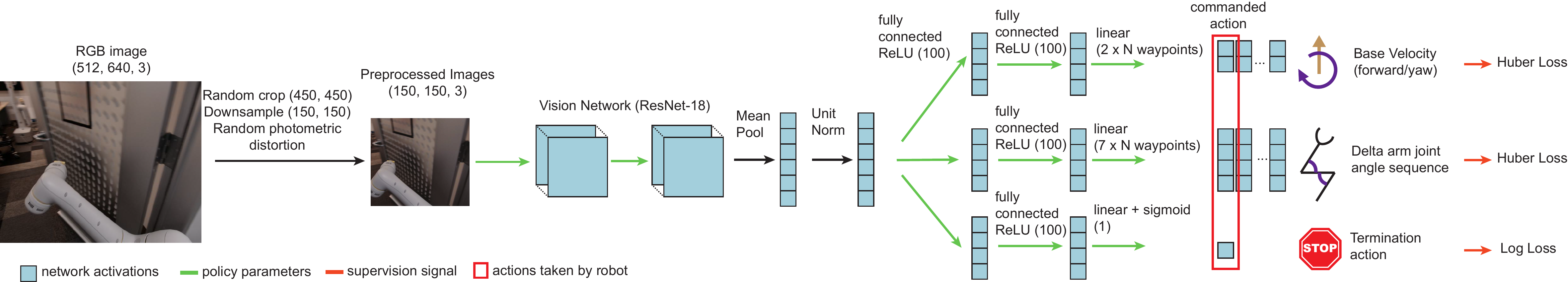}
    \caption{Policy network architecture. The TCL consistency losses are enforced at the image level, feature level at the mean-pool layer, and prediction level (actions).}
    \label{fig:network-architecture}
\end{figure*}

Figure~\ref{fig:network-architecture} displays the network architecture used for all the policies, including the baselines. It uses a similar architecture to \cite{jang2021bcz}, with a ResNet-18 \cite{he2015deep} that projects the mean-pool layer to three ``action heads'': predicted base forward and yaw velocities, predicted arm joints deltas, and whether the policy should terminate the episode instead of moving the robot. Actions are predicted with a 10-step lookahead.

\section{Sampled Images and Domain Adaptation}
\label{appendix:sim2real}

\begin{figure*}[h!]
    \centering
    \includegraphics[width=0.7\textwidth]{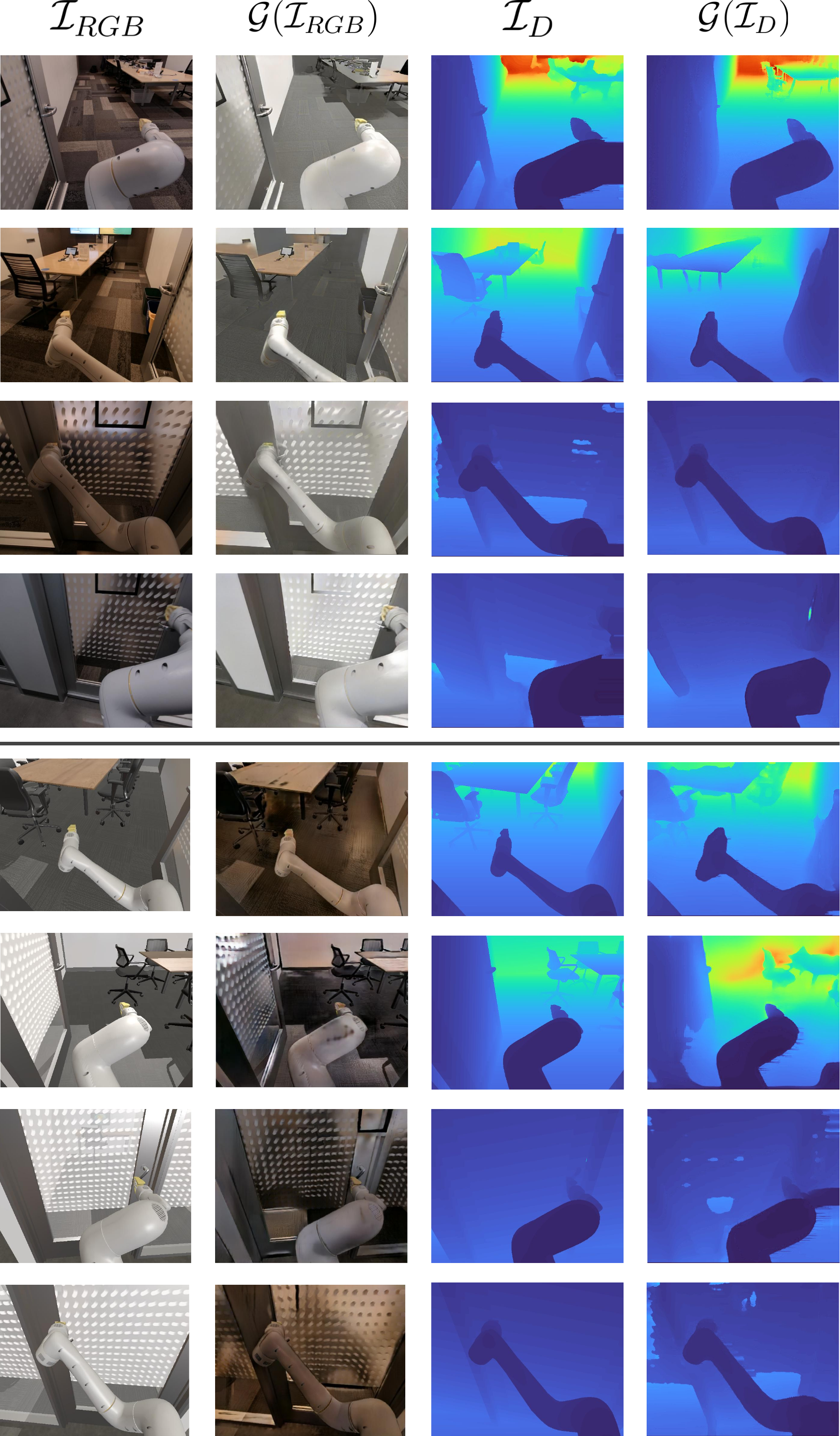}
    \caption{Randomly sampled images from each modality with their domain-adapted counterparts.}
    \label{fig:sim2real-images}
\end{figure*}

Figure \ref{fig:sim2real-images} presents a random sample of simulation and real world images with the domain adaptation adapters $\mathcal{G}$ applied. The top half originate from real world data, while the bottom half originate from simulation. Note the transfer of color tone, lighting, and glass opacity within the RGB images, and note the transfer of noise and glass opacity within the depth images.

\section{Discussion on Simulated vs. Real Evaluations}
As we ultimately care about policy performance in the real world, we need to test our learned models multiple times across a range of scenes to assess generalizability and performance consistency. However, conducting an equivalent set of evaluations in reality vs. simulation can be far more time consuming. As noted in Section \ref{sec:results}, each checkpoints evaluation (requiring 480 runs) takes almost a full day on two robots (including setup time). In contrast the same evaluation in simulation takes approximately <10 minutes using 250 simulated robots. 

For each model training, 100 checkpoints gets exported which takes about <16hr simulation time to evaluate. In contrast, the same evaluation in real world would take 100 days with two robots, and at best 20 days if we use 10 robots (note that we cannot use more than robots in parallel since the total number of rooms is 10). Furthermore, note that the real evaluations require human supervision in case anything goes awry. Without the simulated evaluations, we would also have very low signal regarding which checkpoint to evaluate in reality since simply having a converged BC and TCL loss is not indicative of policy performance. Not only would searching across multiple checkpoints in real be time consuming, but not knowing which checkpoints perform poorly can also be potentially dangerous.

%% file: main.bbl
\begin{thebibliography}{10}\itemsep=-1pt

\bibitem{flowrl}
Artemij Amiranashvili, Alexey Dosovitskiy, Vladlen Koltun, and Thomas Brox.
\newblock Motion perception in reinforcement learning with dynamic objects.
\newblock In {\em "Conference on Robot Learning (CoRL)"}, 2019.

\bibitem{bonin2008visual}
Francisco Bonin-Font, Alberto Ortiz, and Gabriel Oliver.
\newblock Visual navigation for mobile robots: A survey.
\newblock {\em Journal of intelligent and robotic systems}, 53(3):263--296,
  2008.

\bibitem{graspgan}
Konstantinos Bousmalis, Alex Irpan, Paul Wohlhart, Yunfei Bai, Matthew Kelcey,
  Mrinal Kalakrishnan, Laura Downs, Julian Ibarz, Peter Pastor, Kurt Konolige,
  et~al.
\newblock Using simulation and domain adaptation to improve efficiency of deep
  robotic grasping.
\newblock In {\em 2018 IEEE international conference on robotics and automation
  (ICRA)}, pages 4243--4250. IEEE, 2018.

\bibitem{dsn}
Konstantinos Bousmalis, George Trigeorgis, Nathan Silberman, Dilip Krishnan,
  and Dumitru Erhan.
\newblock Domain separation networks.
\newblock {\em Advances in neural information processing systems}, 29:343--351,
  2016.

\bibitem{chen2020simple}
Ting Chen, Simon Kornblith, Mohammad Norouzi, and Geoffrey Hinton.
\newblock A simple framework for contrastive learning of visual
  representations.
\newblock In {\em International conference on machine learning}, pages
  1597--1607. PMLR, 2020.

\bibitem{chen2020exploring}
Xinlei Chen and Kaiming He.
\newblock Exploring simple siamese representation learning, 2020.

\bibitem{autoaugment}
Ekin~D Cubuk, Barret Zoph, Dandelion Mane, Vijay Vasudevan, and Quoc~V Le.
\newblock Autoaugment: Learning augmentation policies from data.
\newblock {\em arXiv preprint arXiv:1805.09501}, 2018.

\bibitem{desouza2002vision}
Guilherme~N DeSouza and Avinash~C Kak.
\newblock Vision for mobile robot navigation: A survey.
\newblock {\em IEEE transactions on pattern analysis and machine intelligence},
  24(2):237--267, 2002.

\bibitem{cutout}
Terrance DeVries and Graham~W Taylor.
\newblock Improved regularization of convolutional neural networks with cutout.
\newblock {\em arXiv preprint arXiv:1708.04552}, 2017.

\bibitem{fang2018multi}
Kuan Fang, Yunfei Bai, Stefan Hinterstoisser, Silvio Savarese, and Mrinal
  Kalakrishnan.
\newblock Multi-task domain adaptation for deep learning of instance grasping
  from simulation.
\newblock In {\em 2018 IEEE International Conference on Robotics and Automation
  (ICRA)}, pages 3516--3523. IEEE, 2018.

\bibitem{DANN}
Yaroslav Ganin, Evgeniya Ustinova, Hana Ajakan, Pascal Germain, Hugo
  Larochelle, Fran{\c{c}}ois Laviolette, Mario Marchand, and Victor Lempitsky.
\newblock Domain-adversarial training of neural networks.
\newblock {\em The journal of machine learning research}, 17(1):2096--2030,
  2016.

\bibitem{GAN}
Ian Goodfellow, Jean Pouget-Abadie, Mehdi Mirza, Bing Xu, David Warde-Farley,
  Sherjil Ozair, Aaron Courville, and Yoshua Bengio.
\newblock Generative adversarial nets.
\newblock {\em Advances in neural information processing systems}, 27, 2014.

\bibitem{gu2017deep}
Shixiang Gu, Ethan Holly, Timothy Lillicrap, and Sergey Levine.
\newblock Deep reinforcement learning for robotic manipulation with
  asynchronous off-policy updates.
\newblock In {\em 2017 IEEE international conference on robotics and automation
  (ICRA)}, pages 3389--3396. IEEE, 2017.

\bibitem{gupta2018robot}
Abhinav Gupta, Adithyavairavan Murali, Dhiraj Gandhi, and Lerrel Pinto.
\newblock Robot learning in homes: Improving generalization and reducing
  dataset bias.
\newblock {\em arXiv preprint arXiv:1807.07049}, 2018.

\bibitem{he2015deep}
Kaiming He, Xiangyu Zhang, Shaoqing Ren, and Jian Sun.
\newblock Deep residual learning for image recognition, 2015.

\bibitem{pba}
Daniel Ho, Eric Liang, Xi Chen, Ion Stoica, and Pieter Abbeel.
\newblock Population based augmentation: Efficient learning of augmentation
  policy schedules.
\newblock In {\em International Conference on Machine Learning}, pages
  2731--2741, 2019.

\bibitem{retinagan}
Daniel Ho, Kanishka Rao, Zhuo Xu, Eric Jang, Mohi Khansari, and Yunfei Bai.
\newblock Retinagan: An object-aware approach to sim-to-real transfer.
\newblock In {\em 2021 IEEE International Conference on Robotics and Automation
  (ICRA)}, pages 10920--10926, 2021.

\bibitem{hodavn2020bop}
Tom{\'a}{\v{s}} Hoda{\v{n}}, Martin Sundermeyer, Bertram Drost, Yann Labb{\'e},
  Eric Brachmann, Frank Michel, Carsten Rother, and Ji{\v{r}}{\'\i} Matas.
\newblock Bop challenge 2020 on 6d object localization.
\newblock In {\em European Conference on Computer Vision}, pages 577--594.
  Springer, 2020.

\bibitem{huber}
Peter~J. Huber.
\newblock {Robust Estimation of a Location Parameter}.
\newblock {\em The Annals of Mathematical Statistics}, 35:73--101, 1964.

\bibitem{henaff2020dataefficient}
Olivier~J. Hénaff, Aravind Srinivas, Jeffrey~De Fauw, Ali Razavi, Carl
  Doersch, S.~M.~Ali Eslami, and Aaron van~den Oord.
\newblock Data-efficient image recognition with contrastive predictive coding,
  2020.

\bibitem{Jain2009BehaviorBasedDO}
Advait Jain and Charles~C. Kemp.
\newblock Behavior-based door opening with equilibrium point control.
\newblock 2009.

\bibitem{rcan}
Stephen James, Paul Wohlhart, Mrinal Kalakrishnan, Dmitry Kalashnikov, Alex
  Irpan, Julian Ibarz, Sergey Levine, Raia Hadsell, and Konstantinos Bousmalis.
\newblock Sim-to-real via sim-to-sim: Data-efficient robotic grasping via
  randomized-to-canonical adaptation networks.
\newblock In {\em Proceedings of the IEEE/CVF Conference on Computer Vision and
  Pattern Recognition}, pages 12627--12637, 2019.

\bibitem{jang2021bcz}
Eric Jang, Alex Irpan, Mohi Khansari, Daniel Kappler, Frederik Ebert, Corey
  Lynch, Sergey Levine, and Chelsea Finn.
\newblock {BC}-z: Zero-shot task generalization with robotic imitation
  learning.
\newblock In {\em 5th Annual Conference on Robot Learning}, 2021.

\bibitem{action_image}
Mohi Khansari, Daniel Kappler, Jianlan Luo, Jeff Bingham, and Mrinal
  Kalakrishnan.
\newblock Action image representation: Learning scalable deep grasping policies
  with zero real world data.
\newblock In {\em 2020 IEEE International Conference on Robotics and Automation
  (ICRA)}, pages 3597--3603, 2020.

\bibitem{levine2016end}
Sergey Levine, Chelsea Finn, Trevor Darrell, and Pieter Abbeel.
\newblock End-to-end training of deep visuomotor policies.
\newblock {\em The Journal of Machine Learning Research}, 17(1):1334--1373,
  2016.

\bibitem{li2019hrl4in}
Chengshu Li, Fei Xia, Roberto Martin-Martin, and Silvio Savarese.
\newblock Hrl4in: Hierarchical reinforcement learning for interactive
  navigation with mobile manipulators, 2019.

\bibitem{retinanet}
Tsung-Yi Lin, Priya Goyal, Ross Girshick, Kaiming He, and Piotr Doll{\'a}r.
\newblock Focal loss for dense object detection.
\newblock In {\em Proceedings of the IEEE international conference on computer
  vision}, pages 2980--2988, 2017.

\bibitem{mundhenk2018improvements}
T~Nathan Mundhenk, Daniel Ho, and Barry~Y Chen.
\newblock Improvements to context based self-supervised learning.
\newblock In {\em Proceedings of the IEEE Conference on Computer Vision and
  Pattern Recognition}, pages 9339--9348, 2018.

\bibitem{noroozi2016unsupervised}
Mehdi Noroozi and Paolo Favaro.
\newblock Unsupervised learning of visual representations by solving jigsaw
  puzzles.
\newblock In {\em European conference on computer vision}, pages 69--84.
  Springer, 2016.

\bibitem{pathak2016context}
Deepak Pathak, Philipp Krahenbuhl, Jeff Donahue, Trevor Darrell, and Alexei~A
  Efros.
\newblock Context encoders: Feature learning by inpainting.
\newblock In {\em Proceedings of the IEEE conference on computer vision and
  pattern recognition}, pages 2536--2544, 2016.

\bibitem{peterson2000high}
L Peterson, David Austin, and Danica Kragic.
\newblock High-level control of a mobile manipulator for door opening.
\newblock In {\em Proceedings. 2000 IEEE/RSJ International Conference on
  Intelligent Robots and Systems (IROS 2000)(Cat. No. 00CH37113)}, volume~3,
  pages 2333--2338. IEEE, 2000.

\bibitem{petrovskaya2007probabilistic}
Anna Petrovskaya and Andrew~Y Ng.
\newblock Probabilistic mobile manipulation in dynamic environments, with
  application to opening doors.
\newblock In {\em IJCAI}, pages 2178--2184, 2007.

\bibitem{rahmatizadeh2018vision}
Rouhollah Rahmatizadeh, Pooya Abolghasemi, Ladislau B{\"o}l{\"o}ni, and Sergey
  Levine.
\newblock Vision-based multi-task manipulation for inexpensive robots using
  end-to-end learning from demonstration.
\newblock In {\em 2018 IEEE international conference on robotics and automation
  (ICRA)}, pages 3758--3765. IEEE, 2018.

\bibitem{rlcyclegan}
Kanishka Rao, Chris Harris, Alex Irpan, Sergey Levine, Julian Ibarz, and Mohi
  Khansari.
\newblock Rl-cyclegan: Reinforcement learning aware simulation-to-real.
\newblock In {\em Proceedings of the IEEE/CVF Conference on Computer Vision and
  Pattern Recognition}, pages 11157--11166, 2020.

\bibitem{ross2011reduction}
St{\'e}phane Ross, Geoffrey Gordon, and Drew Bagnell.
\newblock A reduction of imitation learning and structured prediction to
  no-regret online learning.
\newblock In {\em Proceedings of the fourteenth international conference on
  artificial intelligence and statistics}, pages 627--635. JMLR Workshop and
  Conference Proceedings, 2011.

\bibitem{sadeghi2017cadrl}
Fereshteh Sadeghi and Sergey Levine.
\newblock {CAD2RL}: Real single-image flight without a single real image.
\newblock In {\em Robotics: Science and Systems(RSS)}, 2017.

\bibitem{schmid2008opening}
Andreas~J Schmid, Nicolas Gorges, Dirk Goger, and Heinz Worn.
\newblock Opening a door with a humanoid robot using multi-sensory tactile
  feedback.
\newblock In {\em 2008 IEEE International Conference on Robotics and
  Automation}, pages 285--291. IEEE, 2008.

\bibitem{stuede2019door}
Marvin Stuede, Kathrin Nuelle, Svenja Tappe, and Tobias Ortmaier.
\newblock Door opening and traversal with an industrial cartesian impedance
  controlled mobile robot.
\newblock In {\em 2019 International Conference on Robotics and Automation
  (ICRA)}, pages 966--972. IEEE, 2019.

\bibitem{sun2021fully}
Charles Sun, Jędrzej Orbik, Coline Devin, Brian Yang, Abhishek Gupta, Glen
  Berseth, and Sergey Levine.
\newblock Fully autonomous real-world reinforcement learning for mobile
  manipulation, 2021.

\bibitem{hitnet}
Vladimir Tankovich, Christian Hane, Yinda Zhang, Adarsh Kowdle, Sean Fanello,
  and Sofien Bouaziz.
\newblock Hitnet: Hierarchical iterative tile refinement network for real-time
  stereo matching.
\newblock In {\em Proceedings of the IEEE/CVF Conference on Computer Vision and
  Pattern Recognition}, pages 14362--14372, 2021.

\bibitem{tobin2017domain}
Josh Tobin, Rachel Fong, Alex Ray, Jonas Schneider, Wojciech Zaremba, and
  Pieter Abbeel.
\newblock Domain randomization for transferring deep neural networks from
  simulation to the real world.
\newblock In {\em 2017 IEEE/RSJ international conference on intelligent robots
  and systems (IROS)}, pages 23--30. IEEE, 2017.

\bibitem{wang2020learning}
Cong Wang, Qifeng Zhang, Qiyan Tian, Shuo Li, Xiaohui Wang, David Lane, Yvan
  Petillot, and Sen Wang.
\newblock Learning mobile manipulation through deep reinforcement learning.
\newblock {\em Sensors}, 20(3):939, 2020.

\bibitem{welschehold2017learning}
Tim Welschehold, Christian Dornhege, and Wolfram Burgard.
\newblock Learning mobile manipulation actions from human demonstrations.
\newblock In {\em 2017 IEEE/RSJ International Conference on Intelligent Robots
  and Systems (IROS)}, pages 3196--3201. IEEE, 2017.

\bibitem{wenzel2018modular}
Patrick Wenzel, Qadeer Khan, Daniel Cremers, and Laura Leal-Taix{\'e}.
\newblock Modular vehicle control for transferring semantic information between
  weather conditions using gans.
\newblock In {\em Conference on Robot Learning}, pages 253--269. PMLR, 2018.

\bibitem{xia2021relmogen}
Fei Xia, Chengshu Li, Roberto Martín-Martín, Or Litany, Alexander Toshev, and
  Silvio Savarese.
\newblock Relmogen: Leveraging motion generation in reinforcement learning for
  mobile manipulation, 2021.

\bibitem{yan2018learning}
Xinchen Yan, Jasmined Hsu, Mohi Khansari, Yunfei Bai, Arkanath Pathak, Abhinav
  Gupta, James Davidson, and Honglak Lee.
\newblock Learning 6-dof grasping interaction via deep geometry-aware 3d
  representations.
\newblock In {\em 2018 IEEE International Conference on Robotics and Automation
  (ICRA)}, pages 3766--3773. IEEE, 2018.

\bibitem{yan2019data}
Xinchen Yan, Mohi Khansari, Jasmine Hsu, Yuanzheng Gong, Yunfei Bai, S{\"o}ren
  Pirk, and Honglak Lee.
\newblock Data-efficient learning for sim-to-real robotic grasping using deep
  point cloud prediction networks.
\newblock {\em arXiv preprint arXiv:1906.08989}, 2019.

\bibitem{zhang2017split}
Richard Zhang, Phillip Isola, and Alexei~A Efros.
\newblock Split-brain autoencoders: Unsupervised learning by cross-channel
  prediction.
\newblock In {\em Proceedings of the IEEE Conference on Computer Vision and
  Pattern Recognition}, pages 1058--1067, 2017.

\bibitem{zhang2018deep}
Tianhao Zhang, Zoe McCarthy, Owen Jow, Dennis Lee, Xi Chen, Ken Goldberg, and
  Pieter Abbeel.
\newblock Deep imitation learning for complex manipulation tasks from virtual
  reality teleoperation.
\newblock In {\em 2018 IEEE International Conference on Robotics and Automation
  (ICRA)}, pages 5628--5635. IEEE, 2018.

\bibitem{zhao2019does}
Brady Zhou, Philipp Kr{\"{a}}henb{\"{u}}hl, and Vladlen Koltun.
\newblock Does computer vision matter for action?
\newblock In {\em Science Robotics 22 May 2019: Vol. 4, Issue 30}, 2019.

\bibitem{cyclegan}
Jun-Yan Zhu, Taesung Park, Phillip Isola, and Alexei~A Efros.
\newblock Unpaired image-to-image translation using cycle-consistent
  adversarial networks.
\newblock In {\em Proceedings of the IEEE international conference on computer
  vision}, pages 2223--2232, 2017.

\bibitem{zhu2017target}
Yuke Zhu, Roozbeh Mottaghi, Eric Kolve, Joseph~J Lim, Abhinav Gupta, Li
  Fei-Fei, and Ali Farhadi.
\newblock Target-driven visual navigation in indoor scenes using deep
  reinforcement learning.
\newblock In {\em 2017 IEEE international conference on robotics and automation
  (ICRA)}, pages 3357--3364. IEEE, 2017.

\end{thebibliography}
